\documentclass[11pt,a4paper]{article}
\usepackage[hyperref]{naaclhlt2019}
\usepackage{times}
\usepackage{latexsym}

\usepackage{url}

\usepackage[utf8]{inputenc} % allow utf-8 input
\usepackage[T1]{fontenc}    % use 8-bit T1 fonts
\usepackage{hyperref}       % hyperlinks
\usepackage{booktabs}       % professional-quality tables
\usepackage{amsfonts}       % blackboard math symbols
\usepackage{nicefrac}       % compact symbols for 1/2, etc.
\usepackage{microtype}      % microtypography

\usepackage{amsmath,amsthm, mathtools,verbatim,amssymb,amsfonts,amscd}
\usepackage{multirow, floatrow, blindtext}
\newfloatcommand{capbtabbox}{table}[][\FBwidth]
\usepackage{wrapfig}
\usepackage{adjustbox}
\usepackage{mydefs}
\usepackage{subfigure}

\aclfinalcopy % Uncomment this line for the final submission
\def\aclpaperid{296} %  Enter the acl Paper ID here

\title{Topic-Guided Variational Autoencoders for Text Generation}

\author{Wenlin Wang$^{1}$, Zhe Gan$^{2}$, Hongteng Xu$^{1,3}$, Ruiyi Zhang$^{1}$, Guoyin Wang$^{1}$, \\
\textbf{Dinghan Shen}$^{1}$, \textbf{Changyou Chen}$^{4}$,  \textbf{Lawrence Carin}$^{1}$
  \\
  $^{1}$Duke University, \quad$^{2}$Microsoft Dynamics 365 AI Research, \\\quad$^{3}$Infinia ML, Inc, \quad$^{4}$University at Buffalo\\
  {\tt wenlin.wang@duke.edu } \\
  }

\date{}

\begin{document}
\maketitle
\begin{abstract}
We propose a topic-guided variational autoencoder (TGVAE) model for text generation. 
Distinct from existing variational autoencoder (VAE) based approaches, which assume a simple Gaussian prior for the latent code, our model specifies the prior as a Gaussian mixture model (GMM) parametrized by a neural topic module. 
Each mixture component corresponds to a latent topic, which provides guidance to generate sentences under the topic. 
The neural topic module and the VAE-based neural sequence module in our model are learned jointly.
In particular, a sequence of invertible Householder transformations is applied to endow the approximate posterior of the latent code with high flexibility during model inference.
Experimental results show that our TGVAE outperforms alternative approaches on both unconditional and conditional text generation, which can generate semantically-meaningful sentences with various topics.
\end{abstract}

\section{Introduction}
Text generation plays an important role in various natural language processing (NLP) applications, such as machine translation~\cite{cho2014learning,sutskever2014sequence}, dialogue generation~\cite{li2017adversarial}, and text summarization~\cite{nallapati2016abstractive,rush2015neural}. 
As a competitive solution to this task, the variational autoencoder (VAE)~\cite{kingma2013auto,rezende2014stochastic} has been widely used in text-generation systems~\cite{bowman2015generating, hu2017toward, serban2017hierarchical}.
In particular, VAE defines a generative model that propagates latent codes drawn from a simple prior through a decoder to manifest data samples. 
The generative model is further augmented with an inference network, that feeds observed data samples through an encoder to yield a distribution on the corresponding latent codes. 

Compared with other potential methods, $e.g.$, those based on generative adversarial networks (GANs)~\cite{yu2017seqgan,guo2017long,zhang2017adversarial, zhang2018sequence,chen2018adversarial}, VAE is of particular interest when one desires not only text generation, but also the capacity to infer meaningful latent codes from text. 
Ideally, semantically-meaningful latent codes can provide high-level guidance while generating sentences. 
For example, when generating text, the vocabulary could potentially be narrowed down if the input latent code corresponds to a certain topic (\emph{e.g.}, the word ``military'' is unlikely to appear in a sports-related document). 

However, in practice this desirable property is not fully achieved by existing VAE-based text generative models, because of the following two challenges. 
First, the sentences in documents may associate with different semantic information (\emph{e.g.}, topic, sentiment, etc.) while the latent codes of existing VAE-based text generative models often employ a simple Gaussian prior, which cannot indicate the semantic structure among sentences and may reduce the generative power of the decoder.
Although some variants of VAE try to impose some structure on the latent codes~\cite{jiang2016variational, dilokthanakul2016deep},  they are often designed with pre-defined parameter settings without incorporating semantic meanings into the latent codes, which may lead to over-regularization~\cite{dilokthanakul2016deep}.

\begin{figure*}
	\subfigure[The scheme of our method.]{
		\includegraphics[width=0.6\textwidth]{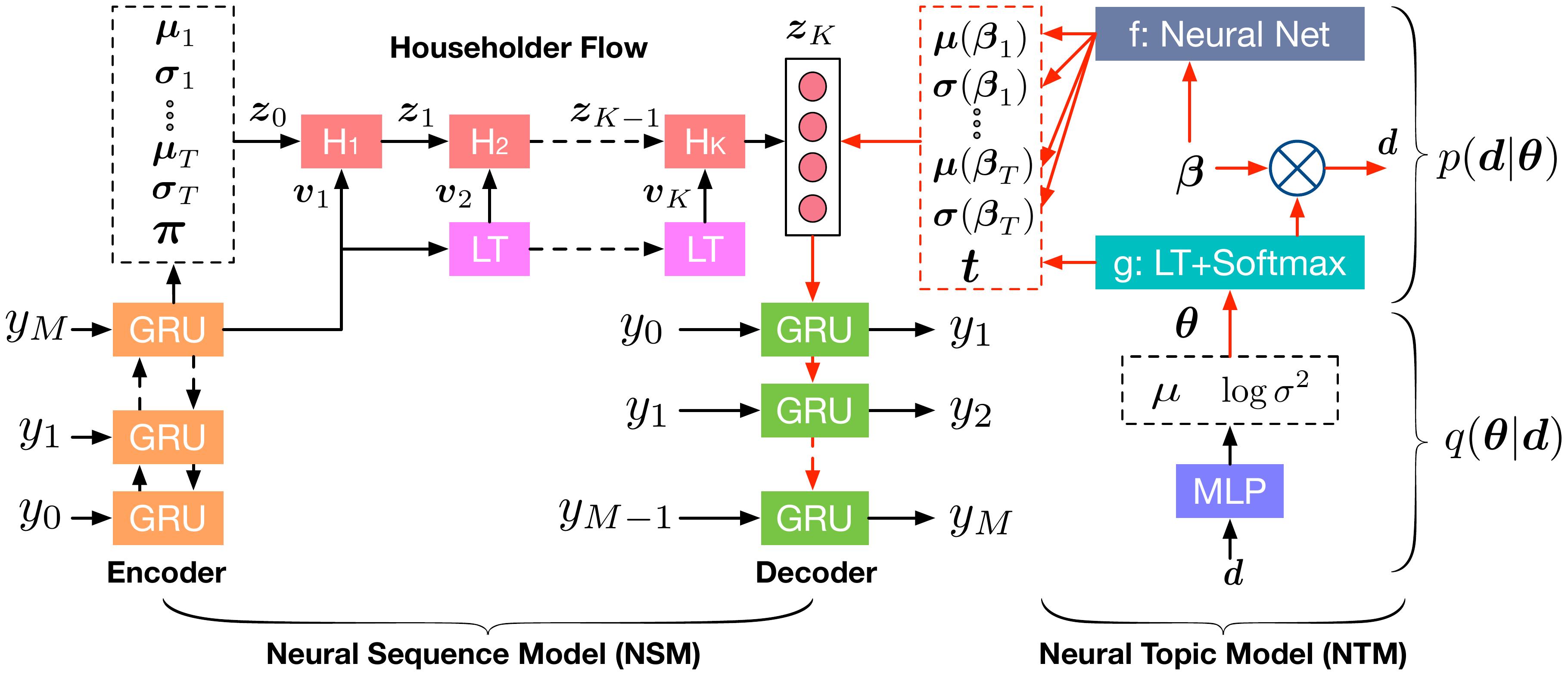}\label{fig: illustration}
	}
	\subfigure[The extension to text summarization.]{
		\includegraphics[width=0.37\textwidth]{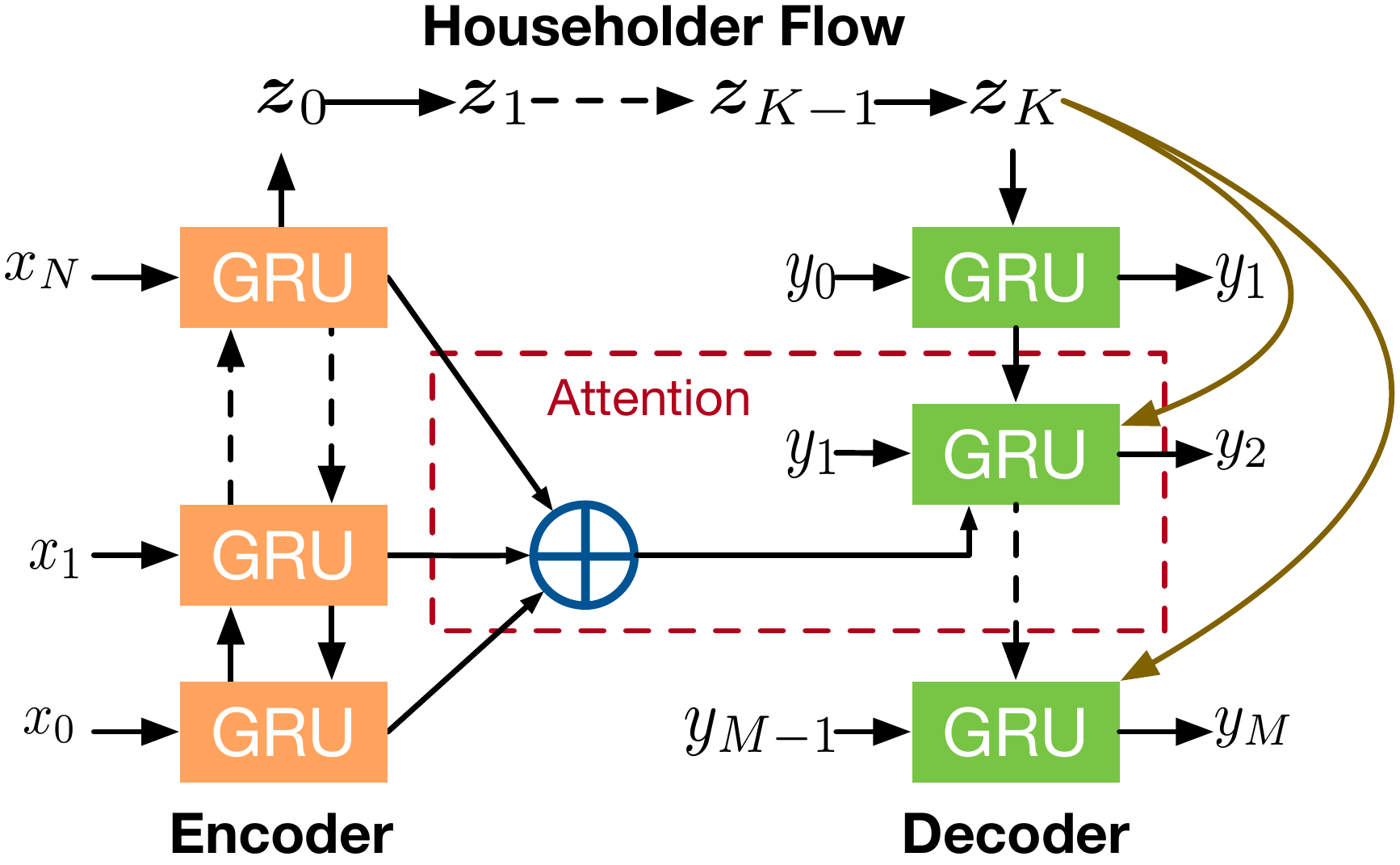}\label{fig: illustration_summarization}
	}
	\caption{\small Illustration of the proposed Topic-Guided Variational Autoencoder (TGVAE) for text generation. (a) For generation (the red arrows), the topics inferred from a neural topic model are used to guide a Gaussian mixture prior of the latent code, which is further fed into the decoder to generate a sentence. For inference (the black arrows), the sentence is encoded into a vector and then propagated through the Householder flow to obtain the approximate posterior. (b) An attention module is further added for text summarization. The same neural topic model is also applied, but omitted here for simplicity of illustration. ``LT'' denotes a linear transformation. }
\end{figure*}

The second issue associated with VAE-based text generation is ``posterior collapse,'' first identified in~\citet{bowman2015generating}. 
With a strong auto-regressive decoder network (\emph{e.g.}, LSTM), the model tends to ignore the information from the latent code and merely depends on previous generated tokens for prediction. 
Several strategies are proposed to mitigate this problem, including making the decoder network less auto-regressive ($i.e.$, using less conditional information while generating each word)~\cite{yang2017improved,shen2017deconvolutional}, or bridging the amortization gap (between the log-likelihood and the ELBO) using semi-amortized inference networks~\cite{kim2018semi}. 
However, these methods mitigate the issue by weakening the conditional dependency on the decoder, which may fail to generate high-quality continuous sentences.  

To overcome the two problems mentioned above, we propose a topic-guided variational autoencoder (TGVAE) model, permitting text generation with designated topic guidance. 
As illustrated in Figure~\ref{fig: illustration}, TGVAE specifies a Gaussian mixture model (GMM) as the prior of the latent code, where each mixture component corresponds to a topic. 
The GMM is learnable based on a neural topic model --- the mean and diagonal covariance of each mixture component is parameterized by the corresponding topic. 
Accordingly, the degree to which each component of the GMM is used to generate the latent code and the corresponding sentence is tied to the usage of the topics. 
In the inference phase, we initialize the latent code from a GMM generated via the encoder, and apply the invertiable Householder transformation~\cite{bischof1994orthogonal,sun1995basis} to derive the latent code with high flexibility and low complexity. 

As shown in Figure~\ref{fig: illustration_summarization}, besides unconditional text generation, the proposed model can be extended for conditional text generation, \textit{i.e.,} abstractive text summarization~\cite{nallapati2016abstractive} with an attention module. 
By injecting the topics learned by our model (semantic information), we are able to make better use of the source document and improve a sequence-to-sequence summarization model~\cite{sutskever2014sequence}.

We highlight the contributions of our model as follows: 
(\emph{i}) A new Topic-Guided VAE (TGVAE) model is proposed for text generation with designated topic guidance. 
(\emph{ii}) For the model inference, Householder flow is introduced to transform a relatively simple mixture distribution into an arbitrarily flexible approximate posterior, achieving powerful approximate posterior inference.
(\emph{iii}) Experiments for both unconditional and conditional text generation demonstrate the effectiveness of the proposed approach.

\section{Model}
The proposed TGVAE, as illustrated in Figure~\ref{fig: illustration}, consists of two modules: a neural topic model (NTM) and  a neural sequence model (NSM). The NTM aims to capture long-range semantic meaning across the document, while the NSM is designed to generate a sentence with designated topic guidance.

\subsection{Neural Topic Model}
Let $\dv\in \mathbb{Z}_+^D$ denote the bag-of-words representation of a document, with $\mathbb{Z}_+$ denoting non-negative integers. $D$ is the vocabulary size, and each element of $\dv$ reflects a count of the number of times the corresponding word occurs in the document. Let $a_n$ represent the topic assignment for word $w_n$. Following~\citet{miao2017discovering}, a Gaussian random vector is passed through  a softmax function to parameterize the multinomial document topic distributions. Specifically, the generative process of the NTM is 
\begin{align}
\thetav \sim \mathcal{N}(0, \Imat)&, \quad  \tv = g(\thetav) \,, \\ 
\quad a_n \sim \text{Discrete}(\tv)&, \quad  w_n \sim \text{Discrete}(\betav_{a_n})\,, \nonumber
\end{align}
where $\mathcal{N}(0, \Imat)$ is an isotropic Gaussian distribution, $g(\cdot)$ is a transformation function that maps sample $\thetav$ to the topic embedding $\tv$, defined here as $g(\thetav) = \mbox{softmax}(\hat{\Wmat}\thetav + \hat{\bv})$, where $\hat{\Wmat}$ and $\hat{\bv}$ are trainable parameters; $\betav_{a_n}$ represents the distribution over words for topic $a_n$; $n\in [1,N_d]$, and $N_d$ is the number of words in the document.
The marginal likelihood for document $\dv$ is:

\vspace{-1em}
{\small
\begin{align}
p(\dv | \betav) &= \int_{\tv}p(\tv)  \sideset{}{_n}\prod \sideset{}{_{a_n}}\sum p(w_n | \betav_{a_n}) p(a_n | \tv) d\tv  \label{eqn:eq1} \\
&= \int_{\tv}p(\tv) \sideset{}{_n}\prod p(w_n | \betav, \tv) d\tv  \nonumber \\
&=  \int_{\tv}p(\tv )  p(\dv | \betav, \tv) d\tv = \int_{\thetav}p(\thetav )  p(\dv | \betav, \thetav) d\thetav \,. \nonumber
\end{align}
}
\vspace{-1em}

The second equation in~(\ref{eqn:eq1}) holds because we can marginalize out the sampled topic words $a_n$ by 

\vspace{-1em}
{\small
\begin{align}
p(w_n | \betav, \tv ) =  \sideset{}{_{a_n}}\sum p(w_n | \betav_{a_n}) p(a_n | \tv)  = \betav\tv \,,
\end{align}
}
\vspace{-1em}

where $\betav = \{\betav_i\}_{i=1}^{T}$ are trainable parameters of the decoder; $T$ is the number of topics and each $\betav_i\in \R^D$ is a topic distribution over words (all elements of $\betav_i$ are nonnegative, and sum to one). 

\subsection{Neural Sequence Model}
Our neural sequence model for text generation is built upon the VAE proposed in~\citet{bowman2015generating}. Specifically, a continuous latent code $\zv$ is first generated from some prior distribution $p(\zv)$, based on which the text sequence $\yv$ is then generated from a conditional distribution $p(\yv|\zv)$ parameterized by a neural network (often called the decoder). Since the model incorporates a latent variable $\zv$ that modulates the entire generation of the sentence, it should be able to capture the high-level source of variation in the data.

\paragraph{Topic-Guided Gaussian Mixture Prior} The aforementioned intuition is hard to be captured by a standard VAE, simply imposing a Gaussian prior on top of $\zv$, since the semantic information associated with a document intrinsically contains different subgroups (such as topics, sentiment, etc.). In our model, we consider incorporating the topic information into latent variables. Our model assumes each $\zv$ is drawn from a \textit{topic-dependent} GMM, that is,
%In contrast to the standard VAE prior that assumes the latent code $\zv$ is drawn from an isotropic Gaussian (\textit{i.e.}, $p(\zv) = \mathcal{N}(0, \mathbf{I})$), the proposed model assumes each $\zv$ is drawn from a \textit{topic-dependent} GMM, that is, 
\begin{align}
p(\zv | \betav, \tv) &= \sideset{}{_{i=1}^T}\sum t_i \mathcal{N}( \muv(\betav_i), \sigmav^2(\betav_i)) ~\nonumber \\
\muv(\betav_i) &= f_{\mu}(\betav_i) ~\nonumber \\
\sigmav^2(\betav_i) &= \text{diag}(\exp{(f_\sigma(\betav_i))}) \,,
\label{eq:meancov}
\end{align} 
where $t_i$ is the usage of topic $i$ in a document and $\betav_i$ is the $i$-th topic distribution over words. Both of them are inherited from the NTM discussed above. Both $f_\mu(\cdot)$ and $f_\sigma(\cdot)$ are implemented as feedforward neural networks, with trainable parameters $\Wmat_\mu$ and $\Wmat_\sigma$, respectively. Compared with a normal GMM prior that sets each mixture component to be $\mathcal{N}(0,\Imat)$, the proposed topic guided GMM prior provides semantic meaning for each mixture component, and hence makes the model more interpretable and controllable for text generation. 

\paragraph{Decoder} The likelihood of a word sequence $\yv = \{y_m\}_{m=1}^M$ conditioned on the latent code $\zv$ is defined as:
\begin{align}
p(\yv|\zv) &= p(y_1|\zv)\sideset{}{_{m=2}^{M}}\prod p(y_m | y_{1:m-1}, \zv) ~\nonumber \\
&=p(y_1|\zv)\sideset{}{_{m=2}^{M}}\prod p(y_m | \hv_m) \,,
\end{align}
where the conditional probability of each word $y_m$ given all the previous words $y_{1:m-1}$ and the latent code $\zv$ is defined through the hidden state $\hv_m$:
$\hv_m = f(\hv_{m-1}, y_{m-1}, \zv)$, where the function $f(\cdot)$ is implemented as a Gated Recurrent Unit (GRU) cell~\cite{cho2014learning} in our experiments.

\section{Inference}
The proposed model (see Figure~\ref{fig: illustration}) takes the bag-of-words as input and embeds a document into a topic vector. The topic vector is then used to reconstruct the bag-of-words input, and the learned topic distribution over words is used to model a topic-dependent prior to generate a sentence in the VAE setup. Specifically, the joint marginal likelihood can be written as:
\begin{align}
p(\yv, \dv | \betav) = \int_\thetav &\int_{\zv} p(\thetav) p(\dv | \betav, \thetav) \nonumber \\
    &\cdot p(\zv | \betav, \thetav) p(\yv|\zv) \,d\thetav d\zv \,. ~\label{jointmarginal}
\end{align}
Since direct optimization of~(\ref{jointmarginal}) is intractable, auto-encoding variational Bayes is employed~\cite{kingma2013auto}. Denote $q(\thetav | \dv)$ and $q(\zv | \yv)$ as the variational distributions for $\thetav$ and $\zv$, respectively. The variational objective function, also called the evidence lower bound (ELBO), is constructed as 

\vspace{-1em}
{\small
\begin{align}
\mathcal{L} =& \underbrace{\mathbb{E}_{q(\thetav | \dv)}\left[ \log p(\dv|\betav, \thetav)\right] - \text{KL}\left( q(\thetav |\dv) || p(\thetav)\right)}_{\text{neural topic model}, \mathcal{L}_t}+ \label{eqn:elbo}  \\
&\underbrace{\mathbb{E}_{q(\zv | \yv)}\left[ \log p(\yv|\zv) \right] - \mathbb{E}_{q(\thetav | \dv)}\left[\text{KL}\left( q(\zv | \yv) || p(\zv|\betav,\thetav)\right)\right]}_{\text{neural sequence model}, \mathcal{L}_s} \nonumber\,.
\end{align}
}
\vspace{-1em}

\noindent By assuming 
\begin{align}
q(\thetav|\dv) = \mathcal{N} (\thetav|g_{\mu}(\dv),\text{diag}(\exp{(g_\sigma(\dv))})), \nonumber
\end{align}
where both $g_\mu(\cdot)$ and $g_\sigma(\cdot)$ are implemented as feed-forward neural networks, the re-parameterization trick~\cite{kingma2013auto} can be applied directly to build an unbiased and low-variance gradient estimator for the $\mathcal{L}_t$ term in (\ref{eqn:elbo}). Below, we discuss in detail how to approximate the $\mathcal{L}_s$ term in (\ref{eqn:elbo}) and infer an arbitrarily complex posterior for $\zv$. Note that $\zv$ is henceforth represented as $\zv_K$ in preparation for the introduction of Householder flows.

\subsection{Householder Flow for Approximate Posterior}
Householder flow~\cite{zhang2017learning, tomczak2016improving} is a volume-preserving normalizing flow~\cite{rezende2015variational}, capable of constructing an arbitrarily complex posterior $q_K(\zv_K|\yv)$ from an initial random variable $\zv_0$ with distribution $q_0$, by composing a sequence of invertible mappings, \emph{i.e.}, $\zv_K = f_K \circ \cdots \circ f_2 \circ f_1 (\zv_0)$. By repeatedly applying the chain rule and using the property of Jacobians of invertible functions, $q_K(\zv_K|\yv)$ is expressed as:

\vspace{-1em}
{\footnotesize
\begin{align}
\log q_K(\zv_K|\yv) = \log q_0 (\zv_0|\yv) - \sideset{}{_{k=1}^K}\sum \log \Big| \det \frac{\partial f_k}{\partial \zv_{k-1}}\Big| \,,
\end{align}
}
\vspace{-1em}

\noindent where $| \det \frac{\partial f_k}{\partial \zv_{k-1}}|$ is the absolute value of the Jacobian determinant. Therefore, the $\mathcal{L}_s$ term in (\ref{eqn:elbo}) may be rewritten as 

\vspace{-1em}
{\small
\begin{align} \label{eqn:normalizing_flow_objective}
&\mathbb{E}_{q_0(\zv_0|\yv)} [\log p(\yv|\zv_K)] +  \sideset{}{_{k=1}^K}\sum \log \Big| \det \frac{\partial f_k}{\partial \zv_{k-1}}\Big| ~\nonumber \\ 
-  &\mathbb{E}_{q(\thetav | \dv)}\left[ \text{KL}(q_0(\zv_0|\yv) || p(\zv_K|\betav,\thetav)) \right] \,.
\end{align}}
\vspace{-1em}

Here $q_0(\zv_0|\yv)$ is also specified as a GMM, \emph{i.e.}, $q_0 (\zv_0 | \yv) = \sum_{i=1}^T \pi_i(\yv) \mathcal{N}(\muv_i(\yv), \sigmav^2_i(\yv))$. As illustrated in Figure~\ref{fig: illustration}, $\yv$ is first represented as a hidden vector $\hv$, by encoding the text sequence with an RNN. Based on this, the mixture probabilities $\piv$, the means and diagonal covariances of all the mixture components are all produced by an encoder network, which is a linear layer with the input $\hv$. 
In (\ref{eqn:normalizing_flow_objective}), the first term can be considered as the reconstruction error, while the remaining two terms act as regularizers, the tractability of which is important for the whole framework.

\paragraph{KL Divergence between two GMMs}
Since both the prior $p(\zv_K|\betav,\thetav)$ and the initial density $q_0(\zv_0|\yv)$ for the posterior are GMMs, the calculation of the third term in (\ref{eqn:normalizing_flow_objective}) requires the KL divergence between two GMMs. Though no closed-form solutions exist, the KL divergence has an explicit upper bound~\cite{dilokthanakul2016deep}, shown in Proposition 1.

\textbf{Proposition 1.} \textit{For any two mixture densities} $p=\sum_{i=1}^{n} \pi_i g_i$ \textit{and} $\hat{p}=\sum_{i=1}^{n} \hat{\pi}_i \hat{g}_i$, \textit{their KL divergence is upper-bounded by}

\vspace{-1em}
{\small
\begin{align}
\text{KL}\left( p || \hat{p} \right) \leq \text{KL}\left(\pi || \hat{\pi} \right) + \sideset{}{_{i=1}^n}\sum \pi_i \text{KL}\left( g_i || \hat{g_i}\right) \,,
\end{align}
}
\vspace{-1em}

\noindent \textit{where equality holds if and only if} $\frac{\pi_ig_i}{\sum_{i=1}^{n}\pi_ig_i} = \frac{\hat{\pi}\hat{g_i}}{\sum_{i=1}^n\hat{\pi}\hat{g_i}}$.

% The proof is an extension of~\cite{do2003fast} and the details are in the Appendix. 
\vspace{0.5em}
\textit{Proof.} With the log-sum inequality

\vspace{-1em}
{\small
\begin{align}
\text{KL}\left( p || \hat{p}\right) &= \int \left(\sum_i \pi_ig_i \right)\log \frac{\sum_i \pi_ig_i}{\sum_i \hat{\pi}\hat{g_i}} \nonumber \\ 
&\leq \int \sum_i \pi_i g_i \log \frac{\pi_i g_i}{\hat{\pi} \hat{g_i}} \nonumber \\
&= \sum_i \pi_i \log \frac{\pi_i}{\hat{\pi}} + \sum_i \pi_i \int g_i \log \frac{g_i}{\hat{g_i}} \nonumber \\
&= \text{KL} (\pi || \hat{\pi}) + \sum_i \pi_i \text{KL} (g_i || \hat{g_i})\,.
\end{align}
}
\vspace{-1em}

Since the KL divergence between two Gaussian distributions has a closed-form expression, the upper bound of the KL divergence between two GMMs can be readily calculated. Accordingly, the third term in (\ref{eqn:normalizing_flow_objective}) is upper bounded as

\vspace{-1em}
{\footnotesize
\begin{align}
&\mathcal{U}_{KL}=\mathbb{E}_{q(\thetav | \dv)}\Bigl[ \text{KL}\left( \piv(\yv) ||\tv  \right) \label{eqn:gmm_kl} \\
&+ \sideset{}{_{i=1}^T}\sum \pi_i (\yv)  \text{KL}\left( \mathcal{N}(\muv_i(\yv), \sigmav_i^2(\yv) || 
\mathcal{N}(\muv(\betav_i), \sigmav^2(\betav_i)) \right) \Bigr], ~\nonumber
\end{align}}
\vspace{-1em}

where the expectation $\mathbb{E}_{q(\thetav | \dv)}[\cdot]$ can be approximated by a sample from $q(\thetav | \dv)$.
\paragraph{Householder Flow}
Householder flow~\cite{tomczak2016improving} is a series of Householder transformations, defined as follows.
For a given vector $\zv_{k-1}$, the reflection hyperplane can be defined by a Householder vector $\vv_t$ that is orthogonal to the hyperplane. The reflection of this point about the hyperplane is 
\begin{align}
\zv_k = \left(\mathbf{I} - 2\frac{\vv_k\vv_k^T}{||\vv_k||^2}\right)\zv_{k-1} = \Hmat_k \zv_{k-1} \,,
~\label{hf}
\end{align}
where $\Hmat_k = \mathbf{I} - 2\frac{\vv_k\vv_k^T}{||\vv_k||^2}$ is called the \textit{Householder matrix}. 
An important property of the \textit{Householder matrix} is that the absolute value of the Jacobian determinant is equal to 1, therefore $\sum_{k=1}^K \log \Big| \det \frac{\partial f_k}{\partial \zv_{k-1}}\Big| = \sum_{k=1}^K \log |\det \Hmat_k| = 0$, significantly simplifying the computation of the lower bound in (\ref{eqn:normalizing_flow_objective}).
For $k=1,\ldots,K$, the vector $\vv_k$ is produced by a linear layer with the input $\vv_{k-1}$, where $\vv_0=\hv$ is the last hidden vector of the encoder RNN that encodes the sentence $\yv$. 

Finally, by combining (\ref{eqn:elbo}), (\ref{eqn:normalizing_flow_objective}) and (\ref{eqn:gmm_kl}), the ELBO can be rewritten as 
\begin{align}
\mathcal{L} \geq \mathcal{L}_t + \mathbb{E}_{q_0(\zv_0|\yv)} [\log p(\yv|\zv_K)] - \mathcal{U}_{KL} \,.
\end{align}

\subsection{Extension to text summarization}
When extending our model to text summarization, we are interested in modeling $p(\yv,\dv|\xv)$, where $(\xv, \yv)$ denotes the document-summary pair, and $\dv$ denotes the bag-of-words of the input document. 
The marginal likelihood can be written as
$p(\yv, \dv |\xv) = \int_\thetav \int_{\zv} p(\thetav) p(\dv | \betav, \thetav)  p(\zv | \betav, \thetav) p(\yv|\xv,\zv) \,d\thetav d\zv$. 
Assume the approximate posterior of $\zv$ is only dependent on $\xv$, \emph{i.e.}, $q(\zv|\xv)$ is proposed as the variational distribution for $\zv$. The ELBO is then constructed as 
\begin{align}
\mathcal{L} = &\mathcal{L}_t 
+ \mathbb{E}_{q(\zv | \xv)}\left[ \log p(\yv|\xv, \zv) \right] \nonumber \\
&- \mathbb{E}_{q(\thetav | \dv)}\left[\text{KL}\left( q(\zv | \xv) || p(\zv|\betav,\thetav)\right)\right] \,,
\end{align}
where $\mathcal{L}_t$ is the same as used in (\ref{eqn:elbo}). The main difference when compared with unconditional text generation lies in the usage of $p(\yv|\xv, \zv)$ and $q(\zv | \xv)$, illustrated in Figure~\ref{fig: illustration_summarization}. The generation of $\yv$ given $\xv$ is not only dependent on a standard Seq2Seq model with attention~\cite{nallapati2016abstractive}, but also affected by $\zv$ (\emph{i.e.}, $\zv_K$), which provides the high-level topic guidance.

\subsection{Diversity Regularizer for NTM}
Redundancy in inferred topics is a common issue existing in general topic models. In order to address this, it is straightforward to regularize the row-wise distance between paired topics to diversify the topics. Following~\citet{xie2015diversifying,miao2017discovering}, we apply a topic diversity regularization while carrying out the inference. 

Specifically, the distance between a pair of topics is measured by their cosine distance $a(\betav_i, \betav_j) = \arccos \left( \frac{|\betav_i\cdot \betav_j|}{\|\betav_i\|_2 \|\betav_j\|_2}\right)$. The mean angle of all pairs of $T$ topics is $\phi = \frac{1}{T^2}\sum_i\sum_j a(\betav_i, \betav_j)$, and the variance is $\nu = \frac{1}{T^2}\sum_i\sum_j(a(\betav_i, \betav_j) - \phi)^2$. Finally, the topic-diversity regularization is defined as $R=\phi - \nu$. 

\section{Related Work}
The VAE was proposed by~\citet{kingma2013auto}, and since then, it has been applied successfully in a variety of applications~\cite{gregor2015draw,kingma2014semi,chen2017continuous, wang2018zero, shen2018nash}. 
Focusing on text generation, the methods in~\citet{miao2017discovering,miao2016neural,srivastava2017autoencoding} represent texts as bag-of-words, and \citet{bowman2015generating} proposed the usage of an RNN as the encoder and decoder, and found some negative results. 
In order to improve the performance, different convolutional designs~\cite{semeniuta2017hybrid,shen2017deconvolutional,yang2017improved} have been proposed. 
A VAE variant was further developed in~\citet{hu2017toward} to control the sentiment and tense of generated sentences. 
Additionally, the VAE has also been considered for conditional text generation tasks, including machine translation~\cite{zhang2016variational}, image captioning~\cite{pu2016variational}, dialogue generation~\cite{serban2017hierarchical,shen2017conditional,zhao2017learning} and text summarization~\cite{li2017deep,miao2016language}. 
In particular, distinct from the above works, we propose the usage of a topic-dependent prior to explicitly incorporate topic guidance into the text-generation framework.

The idea of using learned topics to improve NLP tasks has been explored previously, including methods combining topic and neural language models~\cite{ahn2016neural,dieng2016topicrnn,lau2017topically,mikolov2012context,wang2017topic}, as well as leveraging topic and word embeddings~\cite{liu2015topical, xu2018distilled}.
Distinct from them, we propose the use of topics to guide the prior of a VAE, rather than only the language model (\emph{i.e.}, the decoder in a VAE setup). 
This provides more flexibility in text modeling and also the ability to infer the posterior on latent codes, which could be useful for visualization and downstream tasks. 

%In terms of sequence modeling,
Neural abstractive summarization was pioneered in~\citet{rush2015neural}, and it was followed and extended by~\citet{chopra2016abstractive}. 
Currently the RNN-based encoder-decoder framework with attention~\cite{nallapati2016abstractive,see2017get} remains popular in this area. Attention models typically work as a keyword detector, which is similar to topic modeling in spirit. 
This fact motivated us to extend our topic-guided VAE model to text summarization. 

\begin{table*}[t]
	\begin{center}
		\scalebox{0.8}{
			\begin{tabular}{c|c c|c  c c| c  c c| c c c}
				\specialrule{.1em}{.05em}{.05em} 
				\multirow{2}{*}{\textbf{Dataset}} &
				\multicolumn{2}{c|}{\textbf{Vocabulary}} & \multicolumn{3}{c|}{\textbf{Training}} & \multicolumn{3}{c|}{\textbf{Development}} & \multicolumn{3}{c}{\textbf{Testing}} \\
				& LM & TM & \# Docs & \# Sents & \# Tokens & \# Docs & \# Sents & \# Tokens & \# Docs & \# Sents & \# Tokens \\
				\hline
				\textsc{APNEWS} & $32,400$ & $7,790$& $50K$ & $0.7M$ & $15M$ & $2K$ & $27.4K$ & $0.6M$ & $2K$ & $26.3K$ & $0.6M$ \\ \hline
				\textsc{IMDB} & $34,256$ & $8,713$& $75K$ & $0.9M$ & $20M$ & $12.5K$ & $0.2M$ & $0.3M$ & $12.5K$ & $0.2M$ & $0.3M$ \\ \hline
				\textsc{BNC} & $41,370$ & $9,741$& $15K$ & $0.8M$ & $18M$ & $1K$ & $44K$ & $1M$ & $1K$ & $52K$ & $1M$ \\ 
				\specialrule{.1em}{.05em}{.05em} 
		\end{tabular}}
		\caption{Summary statistics for \textsc{APNEWS}, \textsc{IMDB} and \textsc{BNC}.}
		\label{Table:datasetsStatistics}
	\end{center}
\end{table*}

\section{Experiments}
We evaluate our TGVAE on text generation and text summarization, and interpret its improvements both quantitatively and qualitatively. 
%Implementation details are provided in the Appendix. 

\begin{table*}[t]
	\centering
	\scalebox{0.75}{
		\begin{tabular}{c|c|
				@{\hspace{5pt}}c@{\hspace{5pt}}
				@{\hspace{5pt}}c@{\hspace{5pt}}
				@{\hspace{5pt}}c@{\hspace{5pt}}
				@{\hspace{5pt}}c@{\hspace{5pt}}|
				@{\hspace{5pt}}c@{\hspace{5pt}}
				@{\hspace{5pt}}c@{\hspace{5pt}}
				@{\hspace{5pt}}c@{\hspace{5pt}}
				@{\hspace{5pt}}c@{\hspace{5pt}}|
				@{\hspace{5pt}}c@{\hspace{5pt}}
				@{\hspace{5pt}}c@{\hspace{5pt}}
				@{\hspace{5pt}}c@{\hspace{5pt}}
				@{\hspace{5pt}}c@{\hspace{5pt}}}
			\specialrule{.1em}{.05em}{.05em} 
			\multirow{2}{*}{Metric} &\multirow{2}{*}{Methods} & \multicolumn{4}{c@{\hspace{5pt}}|@{\hspace{5pt}}}{\textsc{APNEWS}} & \multicolumn{4}{c@{\hspace{5pt}}|@{\hspace{5pt}}}{\textsc{IMDB}} & \multicolumn{4}{c}{\textsc{BNC}} \\ \cline{3-14}
			& & B-2 & B-3 & B-4 & B-5 & B-2 & B-3 & B-4 & B-5 & B-2 & B-3 & B-4 & B-5 \\ \hline
			\multirow{8}{*}{\textit{test}-BLEU}
			&VAE                 &  0.564& 0.278 &  0.192&0.122& 0.597& 0.315& 0.219&  0.147&  0.479& 0.266& 0.169&  0.117\\
			&VAE+HF (K=1)         & 0.566 & 0.280 & 0.193& 0.124& 0.593& 0.317& 0.218& 0.148&  0.475& 0.268& 0.165& 0.112\\
			&VAE+HF (K=10)        & 0.570 & 0.279 & 0.195& 0.123& 0.610& 0.322& 0.221& 0.147&  0.483 & 0.270& 0.169& 0.110\\
			&TGVAE (K=0, T=10)   & 0.582 & 0.320 & 0.203& 0.125& 0.627 & 0.362 & 0.223& 0.159&  0.517& 0.282& 0.181& 0.115\\ 
			&TGVAE (K=1, T=10)   & 0.581 & 0.326 & 0.202& 0.124& 0.623 & 0.358 & 0.224& 0.160&  0.519& 0.282&  0.182& 0.118 \\ 
			&TGVAE (K=10, T=10)  & 0.584 & 0.327 & 0.202& 0.126& 0.621& 0.357 & 0.223 &0.159&   0.518& 0.283 & 0.173& 0.119  \\
			&TGVAE (K=10, T=30)  & 0.627 & 0.335 & 0.207& 0.131& \textbf{0.655}& 0.369 & \textbf{0.243} & \textbf{0.165} & 0.528 & \textbf{0.291}& 0.182& 0.119 \\
			&TGVAE (K=10, T=50)  & \textbf{0.629} & \textbf{0.340} & \textbf{0.210}& \textbf{0.132}& 0.652& \textbf{0.372} & 0.239 &0.160 &  \textbf{0.535} & 0.290& \textbf{0.188}&  \textbf{0.120} \\ \hline
			\multirow{8}{*}{\textit{self}-BLEU}
			&VAE                 & 0.866& 0.531&  0.233& -&  0.891& 0.632& 0.275&  -&  0.851& 0.51& 0.163& - \\
			&VAE+HF (K=1)         & 0.865& 0.533&  0.241& -&  0.899& 0.641& 0.278& -& 0.854& 0.515& 0.163& -\\
			&VAE+HF (K=10)        & 0.873& 0.552&  0.219& -&  0.902& 0.648& 0.262& -& 0.854& 0.520& 0.168& -\\
			&TGVAE (K=0, T=10)   &0.847 & 0.499&  0.161& -&  0.878& 0.572& 0.234& -& 0.832& 0.488& 0.160& -\\  
			&TGVAE (K=1, T=10)   &0.847 & 0.495&  0.160& -&  0.871& 0.571& 0.233& -& 0.828& 0.483& 0.150& -\\ 
			&TGVAE (K=10, T=10)  &0.839 & 0.512&  0.172& -&  0.889& 0.577& 0.242& -& 0.829 &0.488& 0.151 &-  \\
			&TGVAE (K=10, T=30)  &0.811 & 0.478&  0.157& -&  0.850& 0.560& 0.231& -& 0.806& 0.473& 0.150 &-  \\
			&TGVAE (K=10, T=50)  &\textbf{0.808} & \textbf{0.476}&  \textbf{0.150}& -&  \textbf{0.842}& \textbf{0.559}& \textbf{0.227}& -& \textbf{0.793}& \textbf{0.469}& \textbf{0.150} &-  \\ \hline
		\end{tabular}
		\caption{\textit{test}-BLEU (higher is better) and \textit{self}-BLEU (lower is better) scores over three corpora.}\label{testBLEU}	
	}
\end{table*}

\begin{table}[t]
	\centering
	\scalebox{0.63}{
		\begin{tabular}{c| c   c  |   c   c | c  c}
			\specialrule{.1em}{.05em}{.05em}
			\multirow{2}{*}{Methods} & \multicolumn{2}{c|}{\textsc{APNEWS}} & \multicolumn{2}{c|}{\textsc{IMDB}} & \multicolumn{2}{c}{\textsc{BNC}} \\ 
			& PPL & KL &  PPL & KL & PPL & KL \\ \hline
			LM  &  62.79&   $-$&   70.38& $-$&  100.07& $-$\\
			LDA+LSTM & 57.05 & $-$ & $69.58$ & $-$ & 96.42 & $-$ \\
			Topic-RNN &56.77 &$-$ &68.74 &$-$ &94.66 &$-$ \\ %~\cite{dieng2016topicrnn}
			TDLM &53.00 &$-$ &63.67 &$-$ &91.42 &$-$ \\ %~\cite{lau2017topically}
			\hline
			VAE &  $\leq$75.89&   1.78&  86.16 & 2.78 & $\leq$105.10  & 0.13  \\
			VAE+HF (K=1) & $\leq$72.99& 1.32&  $\leq$84.06& 1.83& $\leq$105.13 & 0.31 \\
			VAE+HF (K=10) & $\leq$71.60& 0.83&  $\leq$83.67& 1.51&  $\leq$104.82& 0.17 \\
			\hline
			TGVAE (K=0, T=10)  & $\leq$56.12& 2.73&  $\leq$62.99& 3.99 & $\leq$92.32 & 3.40  \\ 
			TGVAE (K=1, T=10)  & $\leq$56.08& 2.70&  $\leq$62.12& 3.86 & $\leq$91.17  & 3.12  \\ 
			TGVAE (K=10, T=10)  & $\leq$55.77& 2.69& $\leq$62.22 &3.94  & $\leq$91.19  &2.99  \\
			TGVAE (K=10, T=30)  & $\leq$51.27& 3.62& $\leq$59.45 &4.62  & $\leq$88.34  &3.82  \\
			TGVAE (K=10, T=50)  & $\leq$\textbf{48.73} & 3.55& $\leq$\textbf{57.11} &5.02 &  $\leq$\textbf{87.86} & 4.57 \\
			\specialrule{.1em}{.05em}{.05em} 
		\end{tabular}
		\caption{Perplexity and averaged KL scores over three corpora. KL in our TGVAE denotes $\mathcal{U}_{KL}$ in Eqn. (\ref{eqn:gmm_kl}).}\label{perplexity}
	}
\end{table}

\subsection{Text Generation} 
\textbf{Dataset} We conduct experiments on three publicly available corpora: APNEWS, IMDB and BNC.\footnote{These three datasets can be downloaded from https://github.com/jhlau/topically-driven-language-model.} APNEWS\footnote{https://www.ap.org/en-gb/} is a collection of Associated Press news articles from 2009 to 2016. 
\textsc{IMDB} is a set of movie reviews collected by~\citet{maas2011learning}, and \textsc{BNC}~\cite{BNCConsortium2007} is the written portion of the British National Corpus, which contains excerpts from journals, books, letters, essays, memoranda, news and other types of text. 
For the three corpora, we tokenize the words and sentences, lowercase all word tokens, and filter out word tokens that occur less than 10 times. For the topic model, we remove stop words in the documents and exclude the top $0.1\%$ most frequent words and also words that appear less than 100 documents. A summary statistics is provided in Table~\ref{Table:datasetsStatistics}.

\noindent\textbf{Evaluation} We first compare the perplexity of our neural sequence model with a variety of baselines. Further, we evaluate BLEU scores on the generated sentences, noted as \textit{test}-BLEU and \textit{self}-BLEU. \textit{test}-BLEU (higher is better) evaluates the quality of generated sentences using a group of real test-set sentences as the reference, and \textit{self}-BLEU (lower is better) mainly measures the diversity of generated samples~\cite{zhu2018texygen}. 

\noindent\textbf{Setup} For the neural topic model (NTM), we consider a 2-layer feed-forward neural network to model $q(\thetav | \dv)$, with 256 hidden units in each layer; ReLU is used as the activation function. The hyper-parameter $\lambda$ for the neural topic model diversity regularizer is fixed to $0.1$ across all the experiments. All the sentences in the paragraph are used to obtain the bag-of-words presentation $\dv$. The maximum number of words in a paragraph is set to 300. For the neural sequence model (NSM), we use bidirectional-GRU as the encoder and a standard GRU as the decoder. The hidden state of our GRU is fixed to $600$ across all the three corpora. For the input sequence, we fix the sequence length to 30. In order to avoid overfitting, dropout with a rate of $0.4$ is used in each GRU layer. 

\noindent\textbf{Baseline} We test the proposed method with different numbers of topics (components in GMM) and different numbers of Householder flows ($i.e.$, $K$), and compare it with six baselines: 
(\textit{i}) a standard language model (LM);
(\textit{ii}) a standard variational RNN auto-encoder (VAE); 
(\textit{iii}) a Gaussian prior-based VAE with Householder Flow (VAE+HF);
(\textit{iv}) a standard LSTM language model with LDA as additional feature (LDA+LSTM);
(\textit{v}) Topic-RNN~\cite{dieng2016topicrnn}, a joint learning framework which learns a topic model and a language model simultaneously;
(\textit{vi}) TDLM~\cite{lau2017topically}, a joint learning framework which learns a convolutional based topic model and a language model simultaneously.

\noindent\textbf{Results} The results in Table~\ref{perplexity} show that the models trained with a VAE and its Householder extension does not outperform a well-optimized language model, and the KL term tends to be annealed with the increase of $K$. 
In comparison, our TGVAE achieves a lower perplexity upper bound, with a relative larger $\mathcal{U}_{KL}$. 
We attribute the improvements to our topic guided GMM model design, which provides additional topical clustering information in the latent space; the Householder flow also boosts the posterior inference for our TGVAE. 
We also observe consistent improvements with the number of topics, which demonstrates the efficiency of our TGVAE. 

To verify the generative power of our TGVAE, we generate samples from our \textit{topic-dependent} prior and compare various methods on the BLEU scores in Table~\ref{testBLEU}. 
With the increase of topic numbers, our TGVAE yields consistently better \textit{self}-BLEU and a boost over \textit{test}-BLEU relative to standard VAE models. 
We also show a group of sampled sentences drawn from a portion of topics in Table~\ref{Table:generateSentences}. 
Our TGVAE is able to generate diverse sentences under topic guidance. 
When generating sentences under a mixture of topics, we draw multiple samples from the GMM and take $\zv$ as the averaged sample.

\begin{table}[t]
	\centering
	\scalebox{0.7}{
	\begin{tabular}{c|  c|    c|  c}
		\specialrule{.1em}{.05em}{.05em}
		\multirow{2}{*}{Methods} & \multicolumn{1}{c|}{\textsc{APNEWS}} & \multicolumn{1}{c|}{\textsc{ IMDB }} & \multicolumn{1}{c}{\textsc{  BNC }} \\ 
		&   T=50 &  T=50 &  T=50 \\ \hline
		LDA~\cite{blei2003latent}       &0.125 & 0.084 & 0.106 \\
		TDLM~\cite{lau2017topically}      & 0.149 & 0.104 &0.102 \\
		Topic-RNN~\cite{dieng2016topicrnn} & 0.134 & 0.103 &0.102 \\
		TGVAE     &  \textbf{0.157}& \textbf{0.105}& \textbf{0.113} \\
		\specialrule{.1em}{.05em}{.05em} 
	\end{tabular} }
	\caption{Topic coherence over \textsc{APNEWS}, \textsc{IMDB} and \textsc{BNC}.}
	\label{Table:coherence}
\end{table} 

Though this paper focuses on generating coherent topic-specific sentences rather than the learned topics themselves, we also evaluate the topic coherence~\cite{lau2017topically} to show the rationality of our joint learning framework. We compute topic coherence using normalized PMI (NPMI). In practice, we average topic coherence over the top 5/10/15/20 topic words. To aggregate topic coherence score, we further average the coherence scores over topics. Results are summarized in Table ~\ref{Table:coherence}.

\begin{table*}[t]
	\begin{center}
		\scalebox{0.6}{
			\begin{tabular}{l | c  l }
				\specialrule{.1em}{.05em}{.05em}
				\textbf{Data} &  \textbf{Topic} & \textbf{Sentences} \\ \hline
				\multirow{9}{*}{\textsc{APNEWS}} & education & $\bullet$ the commission has approved a bill that would make state funding available for the city 's new school .\\
				& animal & $\bullet$the feline did n't survive fence hangars at the lake .\\ 
				& \multirow{2}{*}{crime}& $\bullet$ the jury found the defense was not a <unk> , <unk> 's ruling and that the state 's highest court has been convicted of \\ 
				& & first-degree murder .\\
				& weather & $\bullet$ forecasters say they 're still trying to see the national weather service watch for the latest forecast for friday evening .\\
				&lottory & $\bullet$ she hopes the jackpot now exceeds \$ 9 million .\\ %\cline{2-3}
				& \multirow{2}{*}{education+law} & $\bullet$ an alabama law professor thomas said monday that the state's open court claims it takes an emotional matter about \\ 
				& & issuing child molesters based on religion. \\
				& animal+medicine& $\bullet$ the study says the animal welfare department and others are not sure to make similar cases to the virus in the zoo. \\ \hline
				\multirow{7}{*}{\textsc{IMDB}} & war & $\bullet$ after watching the movie , there is a great documentary about the war in the years of the israeli war .\\
				& children & $\bullet$ the entire animation was great at times as to the readings of disney favorites .\\
				& epsiode & $\bullet$ the show would have warranted for 25 episodes and it does help immediately .\\
				& name & $\bullet$ she steals the other part where norma 's <unk> husband ( crawford ) ( as at his part ,  sh*t for the road ) .\\
				& detective & $\bullet$ holmes shouted just to be as much as the movie 's last scene where there were <unk> pills to nab the <unk> .\\ %\cline{2-3}
				& horror + negative & $\bullet$ the movie about a zombie is the worst movie i have ever seen. \\ 
				& detective + children & $\bullet$ my favorite childhood is that rochester takes the character in jane's way, playing the one with hamlet. \\
				\hline
				\multirow{9}{*}{\textsc{BNC}} & \multirow{2}{*}{medical} & $\bullet$ here mistaking ' causes ' drugs as the problem although both economically ill patients arising from a local job will be \\ 
				& & in traumatic dangers . \\
				& \multirow{2}{*}{education} & $\bullet$ he says the sale is given to five students ' award off : out at a laboratory after the three watts of the hours travelling in\\
				& & and chairman store the bank of the <unk> sutcliffe . \\
				& religion & $\bullet$ schoolchildren will either go or back to church in his place every year in the savoy .\\
				& entertainment & $\bullet$ 100 company and special lace with <unk> garland for tea our garden was filmed after a ceremony  \\
				& IT & $\bullet$ ibm also has shut all the big macs in the 60mhz ncube , represent on the acquisition and mips unix .\\ %\cline{2-3}
				& environment + crime& $\bullet$ the earth's environmental protection agency said that the government was still being shut down by the police. \\
				& education+entertainment& $\bullet$ the school is 55 and hosts one of a musician's theme charities festival.  \\
				\specialrule{.1em}{.05em}{.05em}
		\end{tabular}}
		\caption{Generated sentences from given topics. }\label{Table:generateSentences}
	\end{center}	
\end{table*} 

\begin{table}[t!]
	\begin{center}
			\scalebox{0.63}{
				\begin{tabular}{c |c  c  c  | c  c c}
					\specialrule{.1em}{.05em}{.05em} 
					\multirow{2}{*}{\textbf{Methods}} & \multicolumn{3}{c|}{\textsc{Gigawords}} & \multicolumn{3}{c}{\textsc{DUC-2004}} \\
					& \textbf{RF-1} & \textbf{RF-2} &\textbf{RF-L} & \textbf{RR-1} & \textbf{RR-2} &\textbf{RR-L} \\ \hline
					\text{ABS} & 29.55 & 11.32& 26.42 & 26.55 & 7.06& 22.05\\  %~\cite{rush2015neural}
					\text{ABS+} & 29.78 & 11.89 & 26.97 & 28.18 & 8.49 & 23.81\\  %~\cite{rush2015neural}
					\text{RAS-LSTM} & 32.55 & 14.70 & 30.03 & 28.97 & 8.26 & 24.06 \\ %~\cite{chopra2016abstractive}
					\text{RAS-Elman} & 33.78 & 15.97 & 31.15 & 27.41 & 7.69 & 23.06\\ %~\cite{chopra2016abstractive}
					\text{lvt$2$k-lsent} & 32.67 & 15.59 & 30.64 & 28.35 & 9.46 & 24.59 \\ %~\cite{nallapati2016abstractive}
					\text{lvt$5$k-lsent} & 35.30 & 16.64 & 32.62 & 28.61 & 9.42 & 25.24\\ %~\cite{nallapati2016abstractive}
					\text{ASC+FSC} & 34.17 & 15.94 & 31.92 & 26.73 & 8.39 & 23.88\\ \hline \hline %~\cite{miao2016language}
					\text{Seq2Seq}                &34.03 &15.93 &31.68 &28.39 &9.26 &24.83 \\
					\text{Var-Seq2Seq}            &34.00 &15.97 &31.85 &28.11 &9.24 &24.86 \\
					\text{Var-Seq2Seq-HF (K=1)}    &34.04 &15.98 &31.84 &28.18 &9.27 &24.84 \\
					\text{Var-Seq2Seq-HF (K=10)}   &34.22 &16.10 &32.13 &28.78 &9.11 &24.96 \\
					\text{TGVAE (K=0,  T=10)} &35.34 &16.68 &32.69 &28.99 &9.21 &24.89 \\
					\text{TGVAE (K=1,  T=10)} &35.35 &16.70 &32.64 &29.02 &9.24 &24.93 \\
					\text{TGVAE (K=10, T=10)} &35.40 &16.77 &32.71 &29.07 &9.32 &25.17 \\
					\text{TGVAE (K=10, T=30)} &35.59 &17.18 &32.89 &29.38 &\textbf{9.60} &25.22 \\
					\text{TGVAE (K=10, T=50)} &\textbf{35.63} &\textbf{17.27} &\textbf{33.02} &\textbf{29.65} &9.55 &\textbf{25.38} \\
					\specialrule{.1em}{.05em}{.05em} 
			\end{tabular}}
			\caption{Results on \textsc{Gigawords} and \textsc{DUC-2004}.}
			\label{Table:Rouge}
	\end{center}
	\vspace{-1em}
\end{table}

\subsection{Text Summarization}

\textbf{Dataset} We further test our model for text summarization on two popular datasets. First, we follow the same setup as in~\citet{rush2015neural} and consider the \textsc{Gigawords} corpus\footnote{https://catalog.ldc.upenn.edu/ldc2012t21}, which consists of $3.8$M training pair samples, $190$K validation samples and 1,951 test samples for evaluation. 
An input-summary pair consists of the first sentence and the headline of the source articles. 
We also evaluate various models on the \textsc{DUC-2004} test set\footnote{http://duc.nist.gov/duc2004}, which has $500$ news articles. 
Different from \textsc{Gigawords}, each article in \textsc{DUC-2004} is paired with four expert-generated reference summaries. The length of each summary is limited to $75$ bytes. 

\noindent\textbf{Evaluation} We evaluate the performance of our model with the ROUGE score~\cite{lin2004rouge}, which counts the number of overlapping content between the generated summaries and the reference summaries, \textit{e.g.}, overlapped n-grams. 
Following practice, we use F-measures of ROUGE-1 (RF-1), ROUGE-2 (RF-2) and ROUGE-L (RF-L) for \textsc{Gigawords} and Recall measures of ROUGE-1 (RR-1), ROUGE-2 (RR-2) and ROUGE-L (RR-L) for \textsc{DUC-2004}.

\begin{figure*}[t!]
	\begin{floatrow}
		\capbtabbox{%
			\scalebox{0.58}{
				\begin{tabular}{l} 
					\specialrule{.1em}{.05em}{.05em}
					Sample of Summaries  \\ \specialrule{.05em}{.05em}{.05em}
					%first example
					\textbf{D}: a court here thursday sentenced a \#\#-year-old man to \#\# years in jail after he \\ admitted
					pummelling his baby son to death to silence him while watching television .\\
					\textbf{R}: man who killed baby to hear television better gets \#\# years. \\ 
					\textbf{Seq2Seq}: man sentenced to \#\# years after the son 's death\\
					\textbf{Ours}: a \textcolor{blue}{court} \textcolor{blue}{sentenced} a man \#\# years in \textcolor{blue}{jail}\\ \hline
					%second example
					\textbf{D}: european stock markets advanced strongly thursday on some bargain-hunting \\ and gains by wall street and japanese shares ahead of an expected hike in us \\ interest rates , dealers said \\
					\textbf{R}: european stocks bounce back UNK UNK with closing levels\\ 
					\textbf{Seq2Seq}: european stocks advance ahead of us interest rate hike\\
					\textbf{Ours}: european stocks \textcolor{blue}{rise} on \textcolor{blue}{bargain-hunting}, \textcolor{blue}{dealer} said {\color{red}friday} \\ \hline
					%third example
					\textbf{D}: the democratic people 's republic of korea whitewashed south korea in the women \\'s team semi-finals at the world table tennis championships here on sunday\\
					\textbf{R}: dpr korea sails into women 's team final \\
					\textbf{Seq2Seq}: dpr korea whitewash south korea in women 's team final\\
					\textbf{Ours}: dpr korea {\color{red}beat} south korea in \textcolor{blue}{table tennis} worlds\\
					\specialrule{.1em}{.05em}{.05em}
			\end{tabular} }
		}{%
			\caption{\small Example generated summaries on \textsc{Gigawords}. \textbf{D} is the source article, \textbf{R} means the reference summary, \textbf{Seq2Saeq} represents the summary generated from the Seq2Seq model.}\label{tab:sample}
		}
		\ffigbox[\FBwidth]{%
			%			\hsize0.01\hsize
			\includegraphics[scale=0.33]{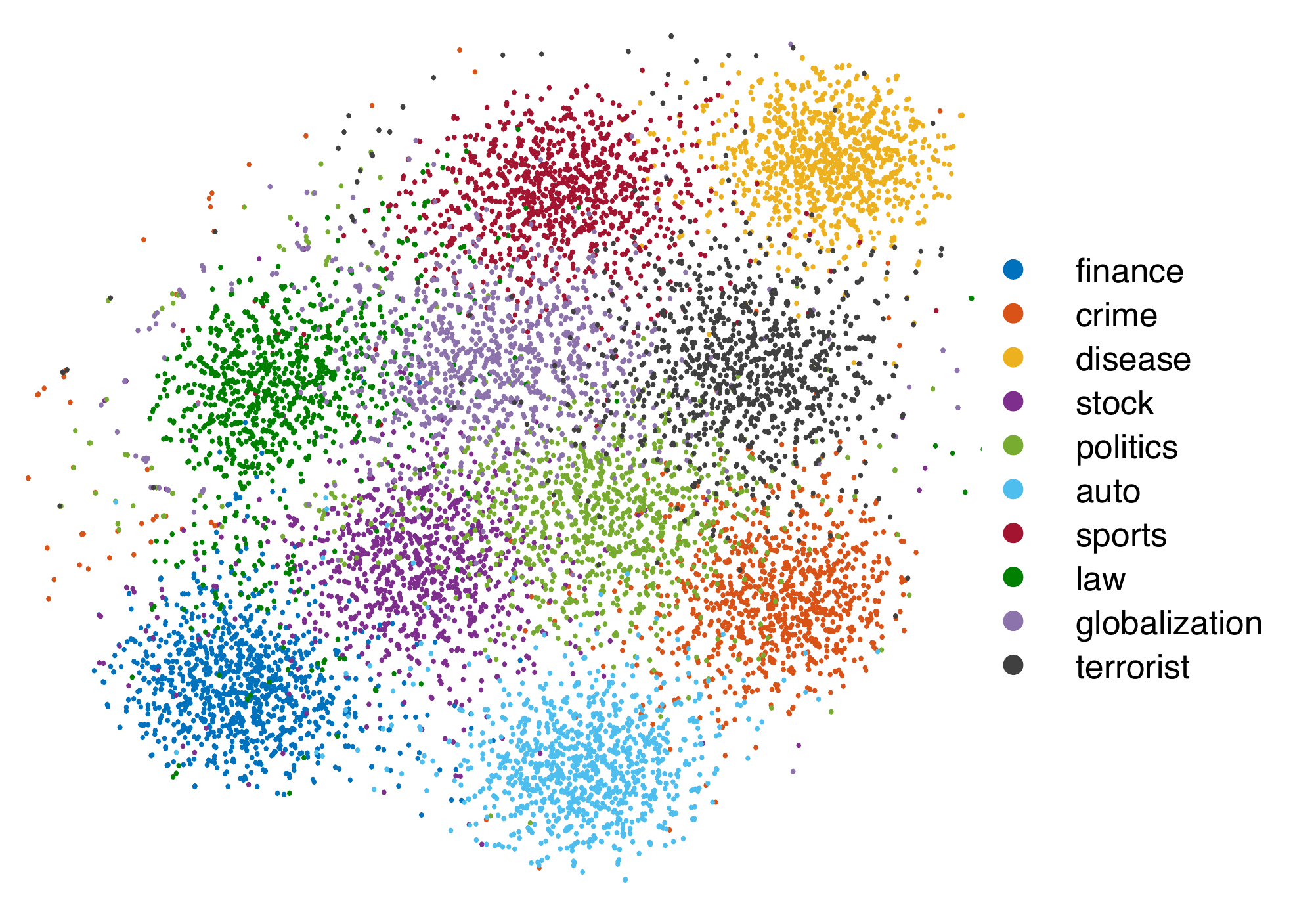}
		}{%
			\caption{\small The t-SNE visualization of $1,000$ samples drawn from the learned topic-guided Gaussian mixture prior.}%
			\label{tsne}
		}
	\end{floatrow}
\end{figure*}

\begin{table*}[t!]
	\begin{center}
		\scalebox{0.58}{
			\begin{tabular}{l | c c c c c c c c c c }
				\specialrule{.1em}{.05em}{.05em} 
				\textbf{Dataset} &  \textbf{education} & \textbf{animal} & \textbf{crime} & \textbf{weather} & \textbf{lottory} &\textbf{terrorism} &  \textbf{law} & \textbf{art} &\textbf{transportation} & \textbf{market} \\ \hline
				\multirow{5}{*}{\textsc{APNEWS}} 
				&students  &animals  &murder       &weather &mega &syria &lawsuit &album & airlines &zacks\\
				&education &dogs     &first-degree &corecasters &lottery &iran &appeals &music & rail &cents \\ 
				&schools   &zoo      &shooting     &winds    &powerball &militants  &justices &film &transit &earnings \\
				&math      &bear     &sentenced    &rain     &gambling &afgan  &constitutional &songs & bridge & revenue\\
				&teachers  &wildlife &gunshot      &snow     &jackpot &korea  &judge &comedy &airport &income \\ \hline
				\multirow{6}{*}{\textsc{IMDB}} & \textbf{war}&\textbf{children} &\textbf{epsiode} &\textbf{name} &\textbf{detective} &\textbf{ethic} &\textbf{action} &\textbf{horror} &\textbf{negative} &\textbf{japanese} \\ \cline{2-11}
				&aircraft  &cinderella &season &crawford &holmes &porn &batman &horror &stupid &miike \\ 
				&president &musical   &episode &stanwyck &poirot &unfunny &king &zombie &horrible &kurosawa \\
				&war       &beatles   &sandler &gable   &christie &sex &chan &werewolf &sucks &sadako \\
				&military  &musicals  &cartoons&powell  &book    &gay &li &candyman &waste &anime \\
				&soldiers  &disney    &jokes   &harlow  &agatha &erotic &ninja &dracula &scary &takashi \\ \hline
				\multirow{6}{*}{\textsc{BNC}} & \textbf{medical} & \textbf{education} & \textbf{religion} & \textbf{entertainment} & \textbf{IT} & \textbf{Law} & \textbf{facilities} & \textbf{crime} & \textbf{sports} & \textbf{environment}\\ \cline{2-11}
				&patients &award       &church &film  &unix &tax &bedrooms &police &cup &nuclear \\
				&gastric  &discipline  &god    &video &corp &coun &hotel &killed &league &emission \\
				&cells    &economic    &art    &album &software &lamont &restaurant &arrested &striker &dioxide \\
				&oesophageal &research &theological &comedy &server &council &rooms &soldiers &season &pollution \\
				&mucosa   &institution &religious &movie &ibm &pensioners &situated &murder &goal &warming \\ \hline
				\multirow{6}{*}{\textsc{Gigawords}} & \textbf{terrorist} & \textbf{crime} & \textbf{finance} & \textbf{sports} & \textbf{law} & \textbf{stock} & \textbf{auto} & \textbf{disease} & \textbf{globalization} & \textbf{politics}\\ \cline{2-11}
				&palestinian &wounding &tael &scored &sentenced &seng &motor &flu &nuclear &republican \\
				&arafat &killed &hk &rebounds &guilty &index &automaker &desease &eu &mccain \\
				&yasser &roadside &gold &points &crimes &prices &toyota &virus &dpark &democrats \\
				&abbas &injuring &cppcc &champion &court &taies &auto &bird &nato &barack \\
				&israeli &crashed &cpc &beats &convicted &stock &ford &health &bilateral &presidential \\ 
				\specialrule{.1em}{.05em}{.05em} 
		\end{tabular}}
		\caption{10 topics learned from our model on \textsc{APNEWS}, \textsc{IMDB}, \textsc{BNC} and \textsc{Gigawords}.}\label{Table:topicwords}
	\end{center}
\end{table*}

\noindent\textbf{Setup} We have a similar data tokenization as we have in text generation. Additionally, for the vocabulary, we count the frequency of words in both the source article the target summary, and maintain the top 30,000 tokens as the source article and target summary vocabulary. For the NTM, we further remove top $0.3\%$ words and infrequent words to get a topic model vocabulary in size of $8000$.  
For the NTM, we follow the same setup as our text generation. In the NSM, we keep using bidirectional-GRU as the encoder and a standard GRU as the decoder. The hidden state is fixed to $400$.
An attention mechanism~\cite{bahdanau2014neural} is applied in our sequence-to-sequence model. 

\noindent\textbf{Baseline} We compare our method with the following alternatives: (\textit{i}) a standard sequence-to-sequence model with attention~\cite{bahdanau2014neural} (Seq2Seq); (\textit{ii}) a model similar to our TGVAE, but without the usage of the topic-dependent prior and Householder flow (Var-Seq2Seq); and (\textit{iii}) a model similar to our TGVAE, but without the usage of the topic dependent prior (Var-Seq2Seq-HF). 

\noindent\textbf{Results} The results in Table~\ref{Table:Rouge} show that our TGVAE achieves better performance than a variety of strong baseline methods on both \textsc{Gigawords} and \textsc{DUC-2004}, demonstrating the practical value of our model. 
It is worthwhile to note that recently several much more complex CNN/RNN architectures have been proposed for abstract text summarization, such as SEASS~\cite{zhou2017selective}, ConvS2S~\cite{gehring2017convolutional}, and Reinforced-ConvS2S~\cite{wang2018reinforced}. In this work, we focus on a relatively simple RNN architecture for fair comparison. In such a way, we are able to conclude that the improvements on the results are mainly from our topic-guided text generation strategy. 
As can be seen, though the Var-Seq2Seq model achieves comparable performance with the standard Seq2Seq model, the usage of Householder flow for more flexible posterior inference boosts the performance. 
Additionally, by combining the proposed topic-dependent prior and Householder flow, we yield further performance improvements, demonstrating the importance of topic guidance for text summarization. 

To demonstrate the readability and diversity of the generated summaries, we present typical examples in Table~\ref{tab:sample}. 
The words in blue are the topic words that appear in the source article but do not exist in the reference, while the words in red are neither in the reference nor in the source article. 
When the topic information is provided, our model is able to generate semantically-meaningful words which may not even exist in the reference summaries and the source articles. 
Additionally, with our topic-guided model, we can always generate a summary with meaningful initial words. 
These phenomena imply that our model supplies more insightful semantic information to improve the quality of generated summaries.

Finally, to demonstrate that our TGVAE learns interpretable topic-dependent GMM priors, we draw multiple samples from each mixture component and visualize them with t-SNE~\cite{maaten2008visualizing}. 
As can be seen from Figure~\ref{tsne}, we have learned a group of separable \textit{topic-dependent} components. 
Each component is clustered and also maintains semantic meaning in the latent space, \textit{e.g.}, the clusters corresponding to the topic ``stock'' and ``finance'' are close to each other, while the clusters for ``finance'' and ``disease'' are far away from each other.  
Additionally, to understand the topic model we have learned, we provide the top 5 words for 10 randomly chosen topics on each dataset (the boldface word is the topic name summarized by us), as shown in Table~\ref{Table:topicwords}.

\section{Conclusion}
A novel text generator is developed, combining a VAE-based neural sequence model with a neural topic model. 
%We impose a hierarchical structure on the latent code of the VAE model with the help of a GMM prior, which provides the VAE model with a topic-specific guidance for text generation. 
The model is an extension of conditional VAEs in the framework of unsupervised learning, in which the topics are extracted from the data with clustering structure rather than predefined labels.
An effective inference method based on Householder flow is designed to encourage the complexity and the diversity of the learned topics. 
Experimental results are encouraging, across multiple NLP tasks.

\clearpage
\bibliography{textVAE}
\bibliographystyle{acl_natbib}

\end{document}